\documentclass{article} 
\usepackage{iclr2027_conference,times}

\usepackage{amsmath,amsfonts,bm}

\def\eqref#1{equation~\ref{#1}}

\def\1{\bm{1}}

\DeclareMathAlphabet{\mathsfit}{\encodingdefault}{\sfdefault}{m}{sl}
\SetMathAlphabet{\mathsfit}{bold}{\encodingdefault}{\sfdefault}{bx}{n}

\usepackage{hyperref}
\usepackage{url}
\usepackage{amsmath, amsfonts, amssymb}
\usepackage{amsthm}
\newtheorem{theorem}{Theorem}

\usepackage{enumitem}
\usepackage{booktabs}
\usepackage{multirow} 
\usepackage{xcolor}
\usepackage{colortbl}
\usepackage[cal=cm]{mathalpha}
\usepackage{mathrsfs}
\usepackage{graphicx}
\title{Generalized Proximal Policy Optimization:\\ Unifying LLM RL Algorithms from the Perspective of Divergence}

\author{Zhicheng Cai, Xinyuan Guo, Hanlin Wu, Weinan Dai, Yuyan Zhou,\\
\textbf{Wei-Ying Ma, Ya-Qin Zhang, Hao Zhou}\thanks{Corresponding author: \texttt{zhouhao@air.tsinghua.edu.cn}}
\\
Institute for AI Industry Research (AIR), Tsinghua University\\
SIA-Lab of Tsinghua AIR and ByteDance Seed
}

\iclrfinalcopy 
\begin{document}

\maketitle

\begin{abstract}
Reinforcement learning has become the foundational paradigm for enhancing the reasoning capabilities of LLMs. 
Yet the dominant approach, Proximal Policy Optimization (PPO), relies on a geometrically degenerate clipping proxy that cannot adapt its resistance to policy shifts, leading to premature clipping and degraded exploration.
Subsequent variants alleviate these issues through fragmented and heuristic modifications. 
We propose \emph{Generalized Proximal Policy Optimization (GPPO)}, a principled and unified framework for understanding and designing LLM RL algorithms through the lens of divergence. 
We establish a generalized $f$-divergence trust-region theory with a monotonic performance improvement guarantee, and show that different $f$-divergences induce distinct geometries and policy update dynamics. 
We further introduce novel estimator topologies that provide more faithful divergence estimates than the conventional single-sample Monte Carlo estimation. 
Under this framework, PPO and recent variants such as DAPO, DPPO, and RIPO emerge as special cases, establishing a principled two-dimensional design space of \textit{Divergence} $\times$ \textit{Estimator}. 
Extensive experiments on eight competition-level math and coding benchmarks show that both dimensions are critical.
The best-performing GPPO instantiation surpasses standard GRPO by up to 50\% and 14\% on math and coding, delivering SOTA performance.
\end{abstract}

\section{Introduction}
Reinforcement learning (RL) has established itself as the primary paradigm for advancing complex reasoning in LLMs~\citep{guo2025deepseek,comanici2025gemini,openai2024learningtoreason}, powering breakthroughs across long-horizon domains such as mathematical reasoning~\citep{shao2024deepseekmath,wang2024math,luo2023wizardmath}, code generation~\citep{dou2024stepcoder,anthropic2024claudecode}, and deep search~\citep{openai2025deepresearch,jin2025searchr1}.

Proximal Policy Optimization (PPO)~\citep{schulman2017proximal} has become a dominant approach for stable and effective RL training, with its “proximal” behavior enforced through clipping. 
Yet beneath this heuristic mechanism lies a geometrically degenerate proxy: PPO implicitly constrains a single-sample Monte Carlo (MC) estimate of the Total Variation (TV) divergence, whose generating function has zero curvature and strictly symmetric bounds. 
This linear geometry cannot adapt its resistance to the magnitude of policy shifts, treating small and large deviations alike. 
This rigidity can lead to premature clipping, distorted trust-region scaling, and exploration collapse~\citep{yu2025dapo,cai2026beyond,qi2026rethinking}.

To alleviate these pathologies, recent variants (e.g., DAPO~\citep{yu2025dapo}, DPPO~\citep{qi2026rethinking}, RIPO~\citep{cai2026beyond}) refine trust-region constraints through relaxing clipping bounds or adjusting Riemannian metrics.
Despite their empirical gains, these methods are largely motivated by specific empirical observations, leading to fragmented modifications of the trust-region constraint. 
Consequently, existing approaches offer no unified, first-principles account of how these seemingly different mechanisms are related, why they improve optimization, or how to systematically design trust-region constraints beyond these isolated heuristics.

We propose Generalized Proximal Policy Optimization (GPPO), a principled framework that unifies the fragmented landscape of LLM RL through the lens of divergence. Rather than introducing another isolated algorithm, GPPO establishes a principled two-dimensional design space of \textit{Divergence} $\times$ \textit{Estimator}, enabling systematic design of proximal policy optimization.

At its core, GPPO establishes a generalized $f$-divergence trust-region theory, showing that policy optimization under $f$-divergence level-set constraints admits a monotonic performance improvement guarantee. 
Crucially, we show that different $f$-divergences induce distinct geometries, providing a principled basis for understanding their different policy update dynamics.

Beyond the choice of divergence, GPPO further generalizes how divergences are estimated, introducing novel estimator topologies that reduce single-sample Monte Carlo variance and provide more faithful divergence estimates. Under this unified lens, PPO and recent variants such as DAPO, DPPO, and RIPO emerge naturally as special cases of GPPO. 

Empirically, our experiments reveal that proximal policy optimization is highly sensitive to both divergence geometry and estimator topology. Compared with the linear and symmetric TV geometry, divergences with non-zero curvature generally yield stronger performance.
Estimator topology is equally important for that Atomic estimation consistently outperforms single-sample MC across all evaluated divergences. 
Notably, KL with Atomic estimation achieves the strongest performance, improving the average score by 50\% on math and 14\% on coding relative to the standard TV-MC formulation. 
These results validate the proposed two-dimensional design space that divergence geometry and estimator topology are both critical to effective RL optimization.

In summary, our contributions are threefold:
\begin{itemize}[leftmargin=*]
\item We propose Generalized Proximal Policy Optimization (GPPO), establishing a two-dimensional design space of \textit{Divergence} $\times$ \textit{Estimator}, which unifies PPO and recent proximal variants, and inspires principled exploration of new proximal policy optimization algorithms.
\item We establish a generalized $f$-divergence trust-region theory with a monotonic performance improvement guarantee, and illustrate how divergence geometry shapes policy update dynamics.
\item We introduce novel estimator topologies that provide more faithful divergence estimates than single-sample Monte Carlo estimation, achieving SOTA performance on math and coding.
\end{itemize}

\section{Preliminary}

\subsection{Monotonic Improvement under Trust Region Constraint}
\cite{schulman2015trust} establishes a trust-region lower bound on the performance improvement between a behavior policy $\mu$ and a target policy $\pi$. Specifically,
\begin{theorem}[Trust-Region Performance Bound]
For behavior policy $\mu$ and target policy $\pi$, the performance improvement satisfies
\begin{equation}
\eta(\pi)-\eta(\mu) \geq \frac{1}{1\!-\!\gamma} \mathbb{E}_{s\sim\rho^\mu, a\sim\mu(\cdot|s)} \left[ r(s,a) A^{\mu}(s,a) \right] - \frac{4\xi\gamma}{(1\!-\!\gamma)^2} D_{\mathrm{TV}}^{\max} (\pi\parallel\mu)^2,
\label{eq:trpo_bound}
\end{equation}
where $\eta(\pi)$ denotes the expected discounted return, $r(s,a)=\frac{\pi(a|s)}{\mu(a|s)}$ is the importance ratio, $A^{\mu}$ is the advantage function under $\mu$, $\xi=\max_{s,a}|A^{\mu}(s,a)|$, and $D_{\mathrm{TV}}^{\max}(\pi\parallel\mu)=\max_s D_{\mathrm{TV}}(\pi(\cdot|s)\parallel\mu(\cdot|s))$ is the maximum state-wise Total Variation (TV) divergence.
\label{thm:trust_region}
\end{theorem}
This bound motivates maximizing the surrogate objective $L_{\mu}(\pi)\!=\!\mathbb{E}_{s\sim\rho^\mu,\,a\sim\mu(\cdot|s)} \left[ r(s,a) A^{\mu}(s,a) \right]$,  while controlling the policy discrepancy measured by $D_{\mathrm{TV}}^{\max}(\pi\parallel\mu)$, providing a formal monotonic improvement guarantee.

This principle underlies Trust Region Policy Optimization (TRPO)~\citep{schulman2015trust}, which maximizes the surrogate objective $\mathcal{L}_{\mu}(\pi)$ subject to a constraint that keeps the new policy within a trust region around the current policy.  Define $\delta > 0$ as the trust region radius, formally, a TV-constrained trust-region formulation can be written as :
\begin{equation}
\begin{aligned}
\max_{\pi}\quad &\mathcal{L}_{\mu}(\pi)\\
\text{s.t.}\quad &D_{\mathrm{TV}}^{\max}(\pi\parallel\mu)\leq\delta.
\label{eq:trpo}
\end{aligned}
\end{equation}
Alternatively, this trust-region constraint can be formulated using the KL divergence, as theoretically justified by Pinsker's inequality: $D_{\mathrm{TV}}(\pi\parallel\mu)^2 \leq \frac{1}{2}D_{\mathrm{KL}}(\pi\parallel\mu)$.

\subsection{Proximal Policy Optimization as a Divergence Proxy}

Despite the theoretical guarantees of TRPO, its constrained second-order optimization incurs substantial computational overhead, limiting its scalability in modern LLM RL. 
Proximal Policy Optimization (PPO)~\citep{schulman2017proximal} addresses this limitation by replacing the second-order optimization with a first-order clipped surrogate objective:
\begin{equation}
\mathcal{J}_{\mathrm{PPO}}(\pi) = \mathbb{E}_{s\sim\rho^\mu,\,a\sim\mu(\cdot|s)} \left[ \min\left( r(s,a)\hat{A}(s,a), \mathrm{clip}\left(r(s,a),1-\epsilon,1+\epsilon\right)\hat{A}(s,a) \right) \right],
\label{eq:ppo_clip}
\end{equation}
where $\hat{A}$ denotes the estimated advantage, typically obtained using GAE~\citep{schulman2015high} in PPO and group-relative advantages in modern LLM RL methods such as GRPO~\citep{guo2025deepseek} and its variants~\citep{yu2025dapo,cai2026beyond}.

The connection between PPO-Clip and the trust-region constraint can be understood through the TV divergence, for each state $s$,
\begin{equation}
D_{\mathrm{TV}}(\pi(\cdot|s)\parallel\mu(\cdot|s)) = \frac{1}{2} \mathbb{E}_{a\sim\mu(\cdot|s)} \left[ \left|r(s,a)-1\right| \right].
\label{eq:tv_ratio}
\end{equation}
Thus, $|r(s,a)-1|$ is a single-sample Monte Carlo estimator of $D_{\mathrm{TV}}(\pi(\cdot|s)\parallel\mu(\cdot|s))$~\citep{qi2026rethinking} (ignore the constant coefficiency). From this perspective, PPO-Clip imposes a per-sample bound (that is \textit{proximal}) on a noisy single-sample proxy of the divergence.

While effective in practice, this proxy has two fundamental limitations. 
First, from a \textit{geometric perspective}, the underlying TV geometry has zero curvature and symmetric penalties, providing no adaptive resistance to different magnitudes of policy shifts. 
Second, from an \textit{estimation perspective}, PPO constrains only a single-sample Monte Carlo estimate rather than the full expectation defining the TV divergence, making the effective trust region sensitive to sampling noise. 
These limitations can manifest as premature clipping, distorted trust-region scaling, and exploration collapse~\citep{yu2025dapo,cai2026beyond,qi2026rethinking}.

\section{Generalized Proximal Policy Optimization}
The preceding analysis reveals that PPO is not merely a clipping heuristic, but a particular instantiation of trust-region policy optimization characterized by a specific divergence and its estimator. 
We introduce \textbf{Generalized Proximal Policy Optimization (GPPO)}, a principled framework that systematically generalizes proximal policy optimization along two complementary dimensions: the \textit{divergence} being constrained and the \textit{estimator} used to compute it.

\subsection{Generalized $f$-Divergence Trust Region}~\label{sec:gtp}
We first generalize the trust-region constraint from Total Variation (TV) to arbitrary $f$-divergences. 
The key theoretical step is to establish a tight quantitative relationship between an arbitrary $f$-divergence and the quadratic TV divergence $D_{\mathrm{TV}}^2$ appearing in the monotonic improvement bound. 
Specifically, Vajda's inequality provides the tightest possible lower bound of an $f$-divergence in terms of TV divergence:
\begin{theorem}[Vajda's Tight Lower Bound]
Let $D_f(P\parallel Q) = \int q(x)f\!\left(\frac{p(x)}{q(x)}\right)\!dx$ be an $f$-divergence generated by a convex function $f$ satisfying $f(1)=0$. 
For $V=D_{\mathrm{TV}}(P\parallel Q)\in[0,1]$, we have
\begin{equation} D_f(P\parallel Q) \geq L_f(V) := \inf_{\beta\in[0,\,1-V]} \left[ \beta f\!\left(1+\frac{V}{\beta}\right) + (1-\beta)f\!\left(1-\frac{V}{1-\beta}\right) \right].
\label{eq:vajda}
\end{equation}
where $\beta\in[0,1-V]$ is the auxiliary probability-mass parameter in the tight binary reduction.
\end{theorem}
Consequently, inverting the tight lower bound yields
$D_{\mathrm{TV}}(P\parallel Q) \leq L_f^{-1}\!\left(D_f(P\parallel Q)\right)$.\\
To establish a tractable quadratic trust-region bound, we seek the largest constant $C_f>0$ such that $D_f(P\parallel Q)\geq C_f D_{\mathrm{TV}}(P\parallel Q)^2$ holds uniformly for all $P,Q$. 
Using Vajda's tight lower bound, this optimal constant is:
\begin{equation}
C_f^{\mathrm{Vajda}} := \inf_{0<V\leq1} \frac{L_f(V)}{V^2},
\label{eq:vajda_constant}
\end{equation}
which establishes the universal quadratic lower bound~\footnote{The detailed derivation is provided in Appendix.}:
\begin{equation}
D_{\mathrm{TV}}(P\parallel Q)^2 \leq \frac{1}{C_f^{\mathrm{Vajda}}} D_f(P\parallel Q).
\label{eq:vajda_quadratic}
\end{equation}
We can now generalize the TV trust-region constraint in Theorem~\ref{thm:trust_region} to an arbitrary $f$-divergence:
\begin{theorem}[Generalized $f$-Divergence Trust-Region Performance Bound]\label{thm:generalized_bound}
Define $D_f^{\max}(\mu\!\parallel\!\pi)\!:=\!\max_s D_f\!\left(\pi(\cdot|s)\parallel\mu(\cdot|s)\right)$.
Then, for any $f$-divergence with $C_f^{\mathrm{Vajda}}>0$,
\begin{equation}
\begin{aligned}
\eta(\pi)-\eta(\mu) &\geq \frac{1}{1-\gamma}\mathcal{L}_\mu(\pi) - \frac{4\xi\gamma}{(1-\gamma)^2} D_{\mathrm{TV}}^{\max}(\pi\parallel\mu)^2 \\
&\geq \frac{1}{1-\gamma}\mathcal{L}_\mu(\pi) - \frac{4\xi\gamma} {C_f^{\mathrm{Vajda}}(1-\gamma)^2} D_f^{\max}(\pi\parallel\mu).
\end{aligned}
\label{eq:generalized_trust_region}
\end{equation}
\vspace{-1em}
\end{theorem}

Thus, constraining any $f$-divergence to a sufficiently small level set guarantees monotonic performance improvement, establishing a unified trust-region framework for TV, Pearson $\chi^2$, KL, Hellinger, and other $f$-divergences.
This naturally motivates the generalized trust-region optimization problem:
\begin{equation}
\begin{aligned}
\max_{\pi}\quad &\mathcal{L}_{\mu}(\pi)\\
\text{s.t.}\quad &D_f^{\max}(\pi\parallel\mu)\leq\delta.
\end{aligned}
\label{eq:gtrpo}
\vspace{-0.5em}
\end{equation}
\paragraph{From $f$-Divergence Trust Regions to Generalized Proximal Policy Optimization.}
Directly solving this generalized constrained optimization problem is computationally prohibitive for large-scale language model policy optimization. 
We therefore retain the first-order proximal paradigm of PPO, while replacing its fixed clipping rule with a clipped proximal region induced by the chosen $f$-divergence. 
Let $\widehat{D}_f$ denote an estimator of the underlying $f$-divergence. 
The corresponding proximal region is defined as the $\delta$-sublevel set of $\widehat{D}_f$:
\begin{equation}
\widehat{\mathcal{R}}_f(\delta)=\left\{r>0:\widehat{D}_f(r)\leq\delta\right\}.
\label{eq:estimated_trust_region}
\end{equation}
This induces the objective of generalized proximal policy optimization:
\vspace{0.3em}
\begin{equation}
\mathcal{J}_{\mathrm{GPPO}}(\pi)=\mathbb{E}_{s\sim\rho^\mu,\,a\sim\mu(\cdot|s)} \left[\min\left(r(s,a)\hat{A}(s,a),\operatorname{clip}_{\widehat{\mathcal{R}}_f(\delta)}(r(s,a))\hat{A}(s,a)\right)\right].
\label{eq:gppo_objective}
\vspace{0.3em}
\end{equation}
Thus the policy ratio is clipped whenever it moves outside the \textit{divergence-induced proximal region} $\widehat{\mathcal{R}}_f(\delta)$, thereby suppressing the gradient associated with the corresponding samples.

\subsection{Divergence Geometry and Policy Update Dynamics}
\label{sec:divergence_geometry}
GPPO in Eq.~\ref{eq:gppo_objective} reduces generalized $f$-divergence trust regions to practical updates.
We now characterize how the geometry of $f$ affects the proximal region and policy update dynamics.
Under our convention in Eq.~\ref{eq:gtrpo}, we constrain $D_f(\pi(\cdot|s)\parallel\mu(\cdot|s)) = \mathbb{E}_{a\sim\mu(\cdot|s)} \left[f(r(s,a))\right]\leq\delta$.
Thus the generating function $f(r)$ determines the shape of the divergence level set and hence the admissible proximal region of policy-ratio, for which its first derivative $f'(r)$ characterizes the marginal divergence cost while its curvature $f''(r)$ governs how that cost grows with the magnitude of a finite ratio shift.
This distinction becomes particularly critical beyond the local regime $r\approx1$.

\vspace{-0.5em}
\paragraph{Local Fisher--Rao Equivalence and Finite-Shift Geometry.}
For smooth $f$-divergences, a Taylor expansion around the reference point $r=1$ gives
\begin{equation}
f(r) = f(1)+f'(1)(r-1) +\frac{1}{2}f''(1)(r-1)^2 +o\!\left((r-1)^2\right).
\label{eq:f_taylor}
\end{equation}
Since $\mathbb{E}_{a\sim\mu}[r-1]=0$ and $f(1)=0$, the linear term vanishes after taking expectation, yielding
\begin{equation}
D_f(\pi\parallel\mu) = \frac{f''(1)}{2} \mathbb{E}_{a\sim\mu}\!\left[(r-1)^2\right] +o\!\left(\|r-1\|^2\right).
\label{eq:f_local_geometry}
\end{equation}
Hence, all twice-differentiable $f$-divergences with $f''(1)>0$ share the same local Fisher--Rao geometry up to a scaling factor.
Their distinction emerges in the finite-shift regime ($r\not\approx1$), where higher-order and asymmetric terms determine how resistance varies with the magnitude and direction of the policy-ratio deviation.

\begin{figure}[t]
  \centering
  \includegraphics[width=0.9\textwidth]{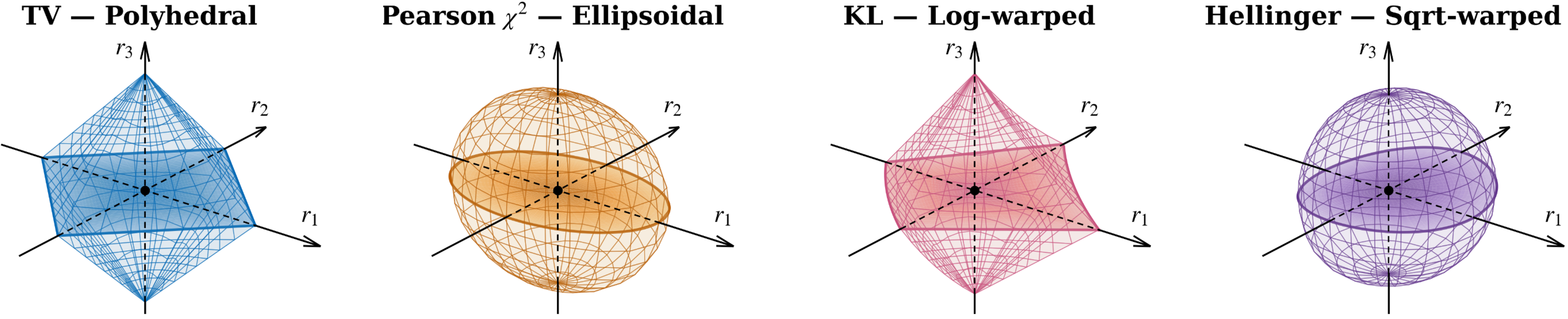}
  \vspace{-1em}
  \caption{Schematic of four $f$-divergence geometries and the corresponding proximal regions they induce, shown for three sampled policy ratios $r=(r_1,r_2,r_3)$ centred at $r=1$. }
  \label{fig:geo}
  \vspace{-1em}
\end{figure}

\paragraph{Zero Curvature of TV and Rigid Clipping.}
For TV divergence, the generating function gives
\vspace{-0.3em}
\begin{equation}
f_{\mathrm{TV}}(r)=\frac{1}{2}|r-1|, \qquad f_{\mathrm{TV}}'(r)=\frac{1}{2}\operatorname{sign}(r-1), \qquad f_{\mathrm{TV}}''(r)=0\quad (r\neq1).
\vspace{-0.3em}
\end{equation}
Thus, TV's polyhedral geometry is piecewise linear with zero curvature almost everywhere, yielding a marginal penalty independent of the magnitude of the ratio deviation $|r-1|$.
PPO-Clip reflects this linear penalty structure through a fixed pointwise clipping threshold, beyond which the corresponding surrogate gradient is suppressed.
Hence, the distinctive limitation of the TV-induced geometry is not the existence of a fixed clipping boundary itself, but the absence of curvature that would otherwise provide magnitude-dependent resistance.
This rigidity provides a geometric explanation for the premature clipping and exploration-related issues discussed in Section~2.2 and explored by subsequent variants~\citep{yu2025dapo,cai2026beyond,qi2026rethinking}.

\textbf{Adaptive Resistance of Smooth Divergences.\quad}
In contrast, smooth $f$-divergences provide adaptive resistance through non-zero curvature.
For Pearson $\chi^2$ divergence,
\vspace{-0.3em}
\begin{equation}
f_{\chi^2}(r)=(r-1)^2, \qquad f_{\chi^2}'(r)=2(r-1), \qquad f_{\chi^2}''(r)=2,
\vspace{-0.3em}
\end{equation}
induces an ellipsoidal (Mahalanobis) geometry where the penalty grows quadratically with the ratio deviation, yielding progressively stronger resistance as $|r-1|$ increases.
For KL divergence,
\vspace{-0.3em}
\begin{equation}
f_{\mathrm{KL}}(r)=-\log r, \qquad f_{\mathrm{KL}}'(r)=-\frac{1}{r}, \qquad f_{\mathrm{KL}}''(r)=\frac{1}{r^2},
\vspace{-0.3em}
\end{equation}
introduces an asymmetric and logarithmically warped geometry where contraction $r\rightarrow0$ is increasingly penalized, while expansion toward large $r$ faces diminishing marginal resistance, thus facilitating probability expansion of low-probability actions and thereby favor exploration. For Hellinger,
\vspace{-0.3em}
\begin{equation}
f_{\mathrm{H}}(r)=(\sqrt{r}-1)^2, \qquad f_{\mathrm{H}}'(r)=1-\frac{1}{\sqrt{r}}, \qquad f_{\mathrm{H}}''(r)=\frac{1}{2r^{3/2}}.
\vspace{-0.3em}
\end{equation}
It induces a spherical geometry under the square-root embedding whose curvature likewise increases toward $r\rightarrow0$ and decreases as $r$ grows, resulting in asymmetric resistance.

Fig.~\ref{fig:geo} visualizes the geometry of above four $f$-divergence. To conclude, \textit{Divergence Geometry} determines the shape of the admissible policy-ratio region and its finite-shift resistance profile, thereby shaping where and how policy updates are constrained.


\subsection{Divergence Estimator Topologies}
\label{sec:estimator_topology}
The proximal region in Eq.~\ref{eq:estimated_trust_region} depends on how the underlying $f$-divergence is estimated. 
Beyond the classical \textit{Single-Sample Monte Carlo} estimator, we propose two alternative topologies, \textit{Atomic} and \textit{Binary}, that more faithfully preserve the distribution-level divergence structure.

\textbf{Single-Sample Monte Carlo Estimation.}
Single-Sample Monte Carlo estimation directly evaluates the divergence integrand at the sampled action:
\begin{equation}
\widehat{D}_f^{\mathrm{MC}}(r(s,a))=f(r(s,a)), \qquad a\sim\mu(\cdot|s).
\label{eq:mc_estimator}
\end{equation}
This induces the proximal region $\widehat{\mathcal{R}}_f^{\mathrm{MC}}(\delta) = \{r > 0 : f(r) \le \delta\}$, suppressing sample gradients whenever $f(r(s,a)) > \delta$.
Although unbiased, $\widehat{D}_f^{\mathrm{MC}}$ relies entirely on single-sample realizations, making its proximal clipping decisions highly susceptible to MC sampling variance.

\textbf{Atomic Estimation.}
To mitigate the sensitivity of MC-estimated constraints to individual sampled outliers, we propose Atomic Estimation, which provides a probability-aware alternative.
Recall that the exact divergence expands over discrete actions as $D_f(\pi\parallel\mu) = \sum_{a} \mu(a|s) f(r(s,a))$.
Atomic estimation imposes the proximal constraint directly on each atomic mass contribution:
\begin{equation}
\widehat{D}_f^{\mathrm{Atom}}(r(s,a)) = \mu(a|s) f(r(s,a)), \qquad a \sim \mu(\cdot|s).
\label{eq:atomic_estimator}
\end{equation}
This induces the probability-aware proximal region $\widehat{\mathcal{R}}_f^{\mathrm{Atom}}(\delta) = \{r > 0 : \mu(a|s) f(r) \le \delta\}$.
Crucially~\footnote{Detailed illustration and complete theoretical analysis are provided in Appendix.}, Atomic estimation distinguishes the magnitude of a ratio deviation from its probability-weighted contribution: a large shift on a low-probability action does not necessarily constitute a large divergence contribution, thereby avoiding premature clipping of rare exploratory actions.
Furthermore, we prove in Appendix that atomic estimation offers an improved bias-variance trade-off than MC, enabling stable and efficient training. 

\textbf{Binary Estimation.}
To capture distribution-level discrepancies beyond the randomly-sampled action, we consider Binary Estimation~\citep{qi2026rethinking}, which collapses the action space into the sampled action and its complement, yielding the binary $f$-divergence estimation:
\begin{equation}
\widehat{D}_f^{\mathrm{Bin}}(r(s,a)) = \mu(a|s)f\left(r(s,a)\right) + \left(1-\mu(a|s)\right)\,f\left(\frac{1-\mu(a|s)r(s,a)}{1-\mu(a|s)}\right), \qquad a \sim \mu(\cdot|s).
\label{eq:binary_estimator}
\end{equation}
This induces the proximal region $\widehat{\mathcal{R}}_f^{\mathrm{Bin}}(\delta) = \{ r(s,a) : \widehat{D}_f^{\mathrm{Bin}}(r(s,a)) \le \delta \}$.
Unlike MC and Atomic estimation, which evaluate only the sampled action contribution, Binary estimation accounts for the complementary probability mass and therefore captures the resulting distribution-level shift.~\footnote{We also explored Top-$K$ estimation since both distributions of $\mu, \pi$ are known. However, it incurs prohibitively high memory overhead ($\mathcal{O}(K)$) and exhibits elevated variance across candidate actions, making the proximal threshold $\delta$ excessively difficult to tune. These empirical findings suggest that overly fine-grained estimation does not necessarily benefit, and can even compromise overall system robustness.}.

To conclude, these estimators provide distinct ways of translating the underlying $f$-divergence geometry into a practical proximal constraint. 


\subsection{Unified View of Existing Proximal Policy Optimization Methods}\label{sec:unified_view}
GPPO provides a principled framework for interpreting existing proximal policy optimization methods within a two-dimensional design space of \textit{Divergence} $f$ $\times$ \textit{Estimator} $\mathcal{T}$. 
Specifically, 

\textbf{PPO}~\citep{schulman2017proximal}, as previously stated, essentially \emph{constrains the TV divergence under single-sample MC estimation}, follower by GRPO~\citep{shao2024deepseekmath,guo2025deepseek}.
GSPO~\citep{zheng2025group} and GMPO~\citep{zhao2025geometric} retain the TV divergence and MC estimator as PPO does, while changing the sampling unit from individual tokens to complete rollouts.

\textbf{DAPO}~\citep{yu2025dapo} adopts an asymmetric clipping mechanism:
\[\mathcal{J}_{\mathrm{DAPO}}(\pi) = \mathbb{E}_{s,a} \left[ \min\left( r(s,a)\hat{A}(s,a), \mathrm{clip}\left(r(s,a),1-\epsilon_\text{low},1+\epsilon_\text{high}\right)\hat{A}(s,a) \right) \right],\]
which corresponds to \emph{constraining the KL divergence under single-sample MC estimation}. According to Eq.~\ref{eq:mc_estimator}, $\widehat{D}_{\mathrm{KL}}^{\mathrm{MC}}(r(s,a))\!=\!-\!\log r$, constraining $-\delta\!\leq\!\log r\!\leq\!\delta$ gives $\widehat{\mathcal{R}}_{\mathrm{KL}}^{\mathrm{MC}}(\delta)\!=\!\{e^{-\delta}\!\leq\!r\!\leq\!e^\delta\}$. Let $\delta=0.23$, thus $0.8\!\leq\!r\!\leq\!1.26$, matching DAPO's asymmetric clipping bounds of $\epsilon_\text{low}$ and $\epsilon_\text{high}$.

\textbf{DPPO}~\citep{qi2026rethinking} constrains $|\pi(a|s)-\mu(a|s)|\leq\delta$, which corresponds exactly to \emph{constraining TV divergence under binary estimation}. 
According to Eq.~\ref{eq:binary_estimator} we obtain, 
\[\widehat{D}_{\mathrm{TV}}^{\mathrm{Bin}}(r(s,a)) = \frac{\mu(a|s)}{2}|r(s,a)\!-\!1| + \frac{\left(1\!-\!\mu(a|s)\right)}{2}\,\left|\frac{1\!-\!\mu(a|s)r(s,a)}{1\!-\!\mu(a|s)}\!-\!1\right| = |\pi(a|s)\!-\!\mu(a|s)|\leq\delta,\] which yields the actual proximal region $\widehat{\mathcal{R}}_{\mathrm{TV}}^{\mathrm{Bin}}(\delta) =\{1-\frac{\delta}{\mu(a|s)} \leq r(s,a)\leq 1+\frac{\delta}{\mu(a|s)}\}$.
Thus, DPPO can be equivalently written as:
\[
\mathcal{J}_{\mathrm{DPPO}}(\pi) = \mathbb{E}_{s,a} \left[ \min\left( r(s,a)\hat{A}(s,a), \mathrm{clip}\left(r(s,a),1-\frac{\delta}{\mu(a|s)},1+\frac{\delta}{\mu(a|s)}\right)\hat{A}(s,a) \right) \right].
\]

\textbf{RIPO}~\citep{cai2026beyond} guarantees isometric policy updates through the clipped objective:
\[
\mathcal{J}_{\mathrm{RIPO}}(\pi) = \mathbb{E}_{s,a} \left[ \min\left( r(s,a)\hat{A}(s,a), \mathrm{clip}\left(r(s,a),1\!-\!\sqrt{\frac{\delta}{\mu(a|s)}},1\!+\!\sqrt{\frac{\delta}{\mu(a|s)}}\right)\hat{A}(s,a) \right) \right].
\]
which essentially \emph{constrains $\chi^2$ divergence under atomic estimation}, since $\widehat{D}_{\mathrm{\chi^2}}^{\mathrm{Atom}}(r(s,a))=\mu(a|s)(r(s,a)-1)^2\leq\delta$ induces exactly the proximal region $\widehat{\mathcal{R}}_{\mathrm{\chi^2}}^{\mathrm{Atom}}(\delta) =\{1\!-\!\sqrt{\frac{\delta}{\mu(a|s)}}\leq r(s,a)\leq1\!+\!\sqrt{\frac{\delta}{\mu(a|s)}}\}$ used by RIPO.

In summary, existing proximal methods correspond to different $(f,\mathcal{T})$ choices, inducing distinct proximal regions for policy updates.
More broadly, GPPO provides a systematic principle for designing new methods: $f$ specifies the desired resistance profile, while $\mathcal{T}$ determines how the resulting divergence constraint is estimated and enforced from sampled data.

\definecolor{c1}{HTML}{d497c3}   
\definecolor{c2}{HTML}{FAC7C8}  
\definecolor{c3}{HTML}{FDEEEE} 
\providecommand{\cI}{\cellcolor{c1}}
\providecommand{\cII}{\cellcolor{c2}}
\providecommand{\cIII}{\cellcolor{c3}}
\providecommand{\est}[2]{#1\textsubscript{\ ({#2})}}
\begin{table*}[!t]
\vspace{-1.5em}
\caption{Results of the different divergences and estimators across various competition-level mathematical reasoning and code generation benchmarks (evaluated by Avg@8). We also compare the performance of other baselines. We highlight each cell as \colorbox{c1}{best}, \colorbox{c2}{second best}, and \colorbox{c3}{third best}.}
\label{tab:main}
\begin{center}
\small
\begin{tabular}{ll|cccc|c}
\toprule
\multicolumn{7}{c}{\textbf{Mathematical Reasoning}} \\
\toprule
Divergence & Estimator & AIME24 & AIME25 & BRUMO25 & SMT25 & Avg. \\
\midrule
\multirow{3}{*}{TV}
& \est{\textit{MC}}{GRPO}    & 34.6 & 20.8 & 33.3 & 21.9 & 27.7 (\ \ +0.00\%) \\
& \est{\textit{Binary}}{DPPO} & 46.7 & \cIII 32.5 & 42.5 & \cII 34.2 & \cIII 39.0 (+40.79\%) \\
& \textit{Atomic} & 45.0 & 28.3 & 40.8 & 30.9 & 36.3 (+31.05\%) \\
\midrule
\multirow{3}{*}{Hellinger}
& \textit{MC }    & 37.9 & 30.8 & 35.4 & 23.1 & 31.8 (+14.80\%) \\
&\textit{Binary} & 42.1 & \cII 32.9 & 37.5 & 29.0 & 35.4 (+27.80\%) \\
& \textit{Atomic} & 42.9 & \cIII 32.5 & \cIII 43.3 & 29.0 & 36.9 (+33.21\%) \\
\midrule
\multirow{3}{*}{$\chi^2$}
& \textit{MC}    & 34.6 & 20.8 & 33.3 & 21.9 & 27.7 (\ \ +0.00\%) \\
& \textit{Binary} & 45.0 & 29.2 & 36.7 & 28.1 & 34.8 (+25.63\%) \\
& \est{\textit{Atomic}}{RIPO} & \cII 47.9 & \cIII 32.5 & \cII 47.5 & \cII 34.2 & \cII 40.5 (+46.21\%) \\
\midrule
\multirow{3}{*}{KL}
& \est{\textit{MC}}{DAPO}    & 45.0 & 27.5 & 37.5 & 28.5 & 34.6 (+24.91\%) \\
& \textit{Binary}& \cIII 47.1 & 29.6 & 42.9 & 29.0 & 37.2 (+34.30\%) \\
& \textit{Atomic} & \cI 48.9 & \cI 33.9 & \cI 48.7 & \cI 34.4 & \cI 41.5 (+49.82\%) \\
\midrule
\multirow{6}{*}{\shortstack[l]{Other\\ Baselines}}
& GSPO  & 42.9  & 28.7  & 36.3  & 32.5  & 35.1 (+26.71\%) \\
& GMPO  & 41.7  & 30.0  & 35.8  & \cIII 33.0 & 35.1 (+26.71\%) \\
& SAPO  & 37.3  & 28.7 & 41.7  & 32.3 & 35.0 (+26.35\%) \\
& CISPO  & 31.7  & 23.8  & 38.3  & 28.0  & 30.5 (+10.11\%) \\
& Clip-Cov  & 36.3  & 22.9  & 36.6  & 30.2  & 31.5 (+13.72\%) \\
& Base Model  & 3.8 & 3.3 & 15.0 & 10.8 & \ \ 8.2 (-70.40\%) \\
\toprule
\multicolumn{7}{c}{\textbf{Code Generation}} \\
\toprule
Divergence & Estimator & Codeforces & CodeContests & TACO & APPS & Avg. \\
\midrule
\multirow{3}{*}{TV}
& \est{\textit{MC}}{GRPO}     & 61.9 & 57.4 & 36.2 & 54.2 & 52.4 (\ \ +0.00\%) \\
& \est{\textit{Binary}}{DPPO} & 63.5 & 58.6 & 37.6 & 55.6 & 53.8 (\ \ +2.67\%) \\
& \textit{Atomic} & 64.2 & 60.5 & 38.1 & 56.7 & 54.9 (\ \ +4.77\%) \\
\midrule
\multirow{3}{*}{Hellinger}
& \textit{MC}     & \cIII 66.4 & 58.2 & 36.3 & 56.8 & 54.4 (\ \ +3.82\%) \\
& \textit{Binary} & 66.3 & 60.2 & 39.2 & 57.7 & 55.9 (\ \ +6.68\%) \\
& \textit{Atomic} & \cII 68.2 & \cII 63.7 & \cII 40.3 & \cII 60.3 & \cII 58.1 (+10.88\%) \\
\midrule
\multirow{3}{*}{$\chi^2$}
& \textit{MC }    & 61.9 & 57.4 & 36.2 & 54.2 & 52.4 (\ \ +0.00\%) \\
& \textit{Binary} & 64.8 & 59.3 & 38.6 & 56.3 & 54.8 (\ \ +4.58\%) \\
& \est{\textit{Atomic}}{RIPO} & \cII 68.2 & \cIII 62.3 & \cIII 39.3 & \cIII 58.9 & \cIII 57.2 (\ \ +9.16\%) \\
\midrule
\multirow{3}{*}{KL}
& \est{\textit{MC}}{DAPO}     & 64.0 & 59.1 & 37.4 & 56.8 & 54.3 (\ \ +3.63\%) \\
& \textit{Binary} & \cIII 66.4 & 61.4 & 38.3 & 57.3 & 55.9 (\ \ +6.68\%) \\
& \textit{Atomic} & \cI 70.7 & \cI 65.1 & \cI 42.3 & \cI 60.8 & \cI 59.7 (+13.93\%) \\
\midrule
\multirow{6}{*}{\shortstack[l]{Other\\ Baselines}}
& GSPO & 63.8 & 57.9 & 36.6 & 56.3  & 53.7 (\ \ +2.48\%) \\
& GMPO & 64.0 & 59.2 & 37.4 & 56.8  & 54.4 (\ \ +3.82\%) \\
& SAPO & 63.7 & 57.0 & 37.5 & 55.9  & 53.5 (\ \ +2.10\%) \\
& CISPO & 61.6 & 54.6 & 35.1 & 53.9  & 51.3 (\ \ -2.10\%) \\
& Clip-Cov & 62.9 & 57.4 & 36.0 & 55.5  & 53.0 (\ \ +1.15\%) \\
& Base Model  & 32.3 & 32.1 & 16.5 & 32.1 & 28.3 (-45.99\%) \\
\bottomrule
\end{tabular}
\end{center}
\vspace{-1.5em}
\end{table*}
\vspace{-1em}
\section{Experiments}
\vspace{-0.3em}
To validate GPPO as a principled framework for proximal policy optimization and to characterize how the constrained divergence and its estimator shape policy update, we conduct extensive experiments across eight competition-level benchmarks spanning both mathematical reasoning and code generation, and benchmark instantiations drawn from the GPPO design space against nine representative RL algorithms.
\vspace{-0.3em}
\subsection{Experimental Setup}
\vspace{-0.3em}
\textbf{Baselines.}
Since GPPO unifies proximal policy optimization along two complementary axes, namely, the constrained \textit{divergence} $f$ and its \textit{estimator} $\mathcal{T}$, we instantiate it across four divergences (TV, Pearson $\chi^2$, Hellinger, and KL) and three estimators (MC, Atomic, and Binary), forming a $4\times3$ design space.
Within this space, several existing methods are recovered as special cases occupying specific $(f,\mathcal{T})$ cells, including GRPO~\citep{shao2024deepseekmath} $(f_\mathrm{TV},\mathcal{T}_\mathrm{MC})$, DAPO~\citep{yu2025dapo} $(f_\mathrm{KL},\mathcal{T}_\mathrm{MC})$, DPPO~\citep{qi2026rethinking} $(f_\mathrm{TV},\mathcal{T}_\mathrm{Bin})$, and RIPO~\citep{cai2026beyond} $(f_{\chi^2},\mathcal{T}_\mathrm{Atom})$.

Rather than tuning the trust-region radius $\delta$ directly, we calibrate it per divergence by matching the contraction-side clip of its MC proximal region $\widehat{\mathcal{R}}_f^{\mathrm{MC}}(\delta)=\{r>0:f(r)\le\delta\}$ to standard PPO-Clip, isolating divergence geometry from trust-region scale: fixing the lower bound to $r_{\min}=0.8$ directly gives $\delta=f(0.8)$.
Accordingly, $\delta_{\mathrm{TV}}=0.2,\ \delta_{\chi^2}=0.04,\ \delta_{\mathrm{H}}\approx0.01$, and  $\delta_{\mathrm{KL}}\approx0.22$.
While the upper bound $r_{\max}$ follows from $f(r_{\max})=\delta$. 
The Atomic and Binary estimators directly reuse this $\delta$, so that all estimators of a divergence share a common radius while their realized bounds adapt to the reference probability $\mu(a\mid s)$.

We further compare against other RL algorithms, including GSPO~\citep{zheng2025group} and GMPO~\citep{zhao2025geometric}, which apply the TV--MC constraint at the full-rollout rather than per-token level; 
SAPO~\citep{gao2025soft} and CISPO~\citep{chen2025minimax}, which preserve otherwise-clipped gradients via soft gating and importance-weight clipping; and Clip-Cov~\citep{cui2025entropy}, which clips high-covariance tokens.
All following their default hyper-parameter settings.

Following common practice~\citep{yang2025dcpo,ye2020mastering}, we apply dual clipping to the importance ratio with lower and upper bounds of $0.5$ and $10$, and remove the KL penalty term for all methods~\citep{yu2025dapo,yang2025dcpo}.
All algorithms adopt the GRPO-style group-relative advantage estimator, and trained on Qwen3-8B-Base~\citep{yang2025qwen3}.

\textbf{Training.}
We evaluate the improvement induced by each RL algorithm on two complementary domains, mathematical reasoning and code generation, both of which admit verifiable rewards and are widely regarded as strong indicators of general reasoning ability.
For mathematical reasoning we train on DAPO-Math-17k~\citep{yu2025dapo}, and for code generation on Eurus-Code-25k~\citep{cui2025process}.
For each query we generate 8 rollouts (namely, the group size is 8) with a maximum response length of 20k tokens for thorough reasoning.
In each RL iteration, the behavior policy $\pi_{\theta_\text{old}}$ totally generates 1024 rollouts with a train batch size of 128, and the current policy $\pi_\theta$ is updated 8 times with a mini-batch size of 16.
Rewards are set to $1$ for correct responses and $0$ otherwise.
We uniformly adopt token-mean loss aggregation and optimize all models to convergence using AdamW~\citep{loshchilov2017decoupled} with a constant learning rate of $1\times10^{-6}$, training for 400 steps on mathematical reasoning and 600 steps on code generation.
All experiments are conducted on the VeRL~\citep{sheng2025hybridflow} framework using 8$\times$H20 GPUs.

\textbf{Evaluation.}
We comprehensively evaluate the effectiveness of these algorithms on eight competition-level benchmarks without contamination, spanning the two domains.
For mathematical reasoning we use AIME24, AIME25, BRUMO25, and SMT25~\citep{balunovic2025matharena}; for code generation we use Codeforces, CodeContests, TACO, and APPS~\citep{cui2025process}.
We repeat the evaluation set for 8 times and report $\text{avg}@8$ metric for results stability. 

\begin{figure}[t]
  \centering
  \includegraphics[width=0.98\textwidth]{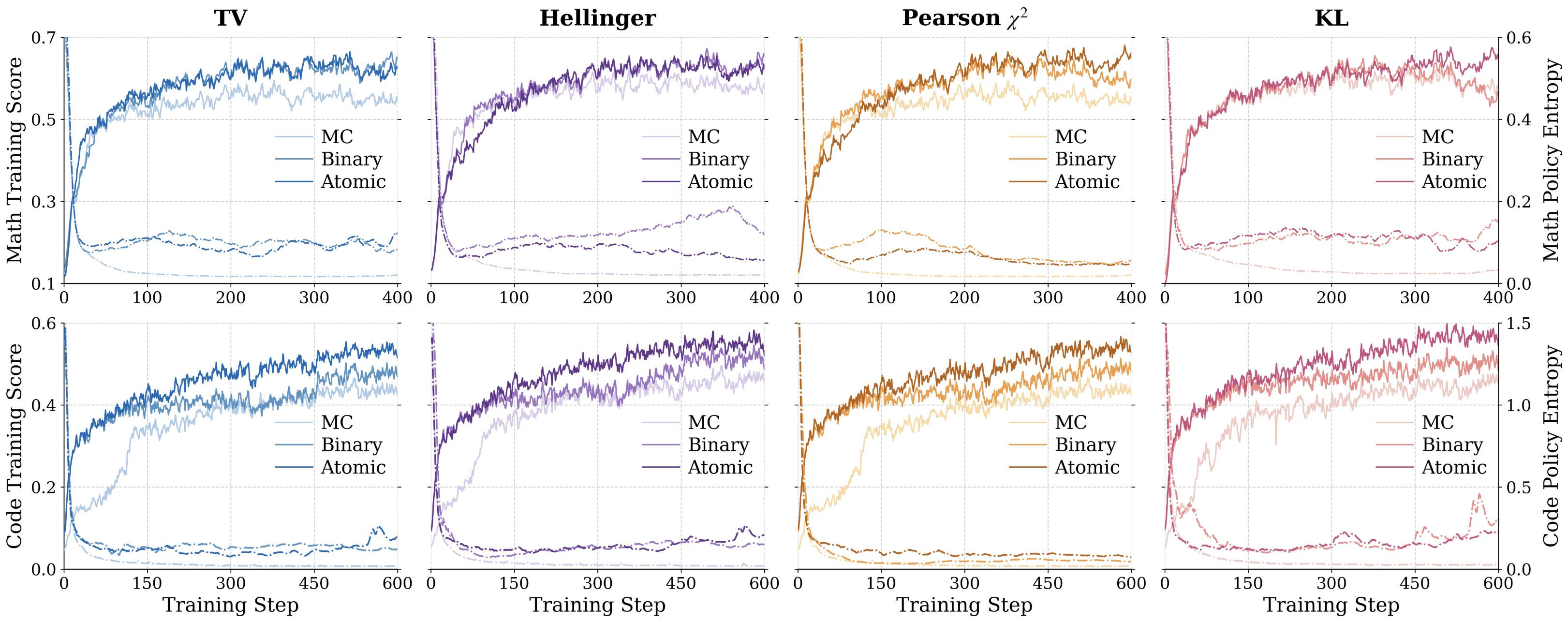}
  \vspace{-1em}
  \caption{Training dynamics of different GPPO instantiations on Math and Code. Solid lines represent the training score, while dash-dotted lines represent the policy entropy.}
  \label{fig:dynamics}
  \vspace{-1em}
\end{figure}

\subsection{Main Results}
Table~\ref{tab:main} summarizes the main experimental results across eight competition-level benchmarks.
Fig~\ref{fig:dynamics} visualized the training dynamics of different GPPO instantiations on math and coding.

\textbf{GPPO establishes a new state-of-the-art across reasoning tasks.}
As highlighted in Table~\ref{tab:main}, our top-performing GPPO instantiation (KL divergence under Atomic estimation) achieves an average of $41.5$ on mathematical reasoning and $59.7$ on code generation, drastically improving upon the standard TV--MC baseline (GRPO) by $+49.8\%$ and $+13.9\%$, respectively, outperforming all existing RL algorithms.
Beyond this optimal configuration, GPPO instantiations consistently dominate the design space, every divergence equipped with an Atomic or Binary estimator outperforms the strongest baseline (GMPO). 
These gains are uniform rather than dataset-specific, confirming that systematic exploration of the \textit{Divergence} $\times$ \textit{Estimator} interplay provides a fundamentally superior foundation for policy optimization than ad-hoc clipping heuristics.

\textbf{Divergence geometry shapes policy optimization dynamics.}
Under single-sample MC, KL improves over TV by $24.9\%$ on math and $3.6\%$ on code, while Hellinger yields gains of $14.8\%$ and $3.8\%$, both enhancing the entropy level of TV. 
TV--MC and $\chi^2$--MC share the same results for that our matched calibration of single-sample MC coincidentally assigns them the same symmetric interval $r\in[0.8,1.2]$, their geometric differences become contributing when estimated more faithfully with Atomic or Binary.
Under Atomic estimation, KL surpasses TV by $+14.3\%$ and $+8.7\%$ on math and code.
Binary reverses this on math but yields only modest gains for non-TV divergences on code, showing that divergence geometry is filtered through estimator topology rather than universally ranked.
One possible explanation is that TV--Binary directly constrains the absolute mass change $|\pi(a\mid s)-\mu(a\mid s)|\leq\delta$, whereas curved divergences convert the same radius into a $\mu(a\mid s)$-dependent ratio penalty that can be tighter for rare actions. This could favor TV when useful mathematical updates require reallocating mass among a few low-probability reasoning branches, while coding's larger set of valid implementations may attenuate the difference.

\textbf{Estimator topology determines how this geometry is enforced under sampling.}
Across four divergences, both proposed probability-aware estimators improve substantially over MC, while Atomic ranks first in every column of the code table and achieves the best math average.
Averaged over the four divergences, Binary improves over MC by $21.3\%$ on math and $3.2\%$ on code, whereas Atomic yields larger gains of $28.3\%$ and $7.7\%$, respectively.
As Figure~\ref{fig:dynamics} shows, the entropy of MC collapse rapidly toward zero with the score curves plateau early at comparatively low levels.
In contrast, Atomic and Binary maintain moderate range of entropy and continue improving over a longer horizon.
This phenomenon follows directly from the estimator: Atomic avoids clipping large but low-probability exploratory updates by weighting with reference probability, and Binary instead accounts for the complementary probability mass.

\vspace{-0.5em}
\section{Conclusion}
\vspace{-0.5em}
This paper introduces GPPO, a principled framework that unifies proximal policy optimization through a two-dimensional design space
of divergence geometry and estimator topology, supported by a generalized $f$-divergence trust-region theorem. 
Across eight mathematical reasoning and code generation benchmarks, GPPO substantially outperforms standard PPO, demonstrating that both dimensions are critical for effective LLM reinforcement learning. 
Ultimately, GPPO provides a systematic foundation for understanding and designing future LLM RL algorithms.


\clearpage



\bibliography{iclr2027_conference}
\bibliographystyle{iclr2027_conference}

\clearpage
\appendix
\section{Derivation for Theorem~2}
\paragraph{Derivation of Vajda's Tight Lower Bound.}

We derive the tight lower bound by reducing the original probability space to a binary partition induced by the total variation distance. Let
\[
V=D_{\mathrm{TV}}(P\parallel Q)
=\frac{1}{2}\int_{\mathcal{X}}|p(x)-q(x)|\,dx.
\]
Consider the Hahn decomposition induced by the signed measure $P-Q$,
with
\[
A:=\{x\in\mathcal{X}:p(x)\geq q(x)\}.
\]
and let $A^c=\mathcal{X}\setminus A$ denote its complement. By construction,
$p(x)-q(x)\geq 0$ on $A$ and $p(x)-q(x)<0$ on $A^c$. Therefore,
\begin{align}
2V
&=\int_A(p(x)-q(x))\,dx
+\int_{A^c}(q(x)-p(x))\,dx \notag\\
&=P(A)-Q(A)+Q(A^c)-P(A^c).
\label{eq:vajda_tv_partition}
\end{align}
Since $A$ and $A^c$ form a partition of $\mathcal{X}$,
\[
P(A^c)=1-P(A),
\qquad
Q(A^c)=1-Q(A).
\]
It follows that
\begin{align}
Q(A^c)-P(A^c)
&=(1-Q(A))-(1-P(A)) \notag\\
&=P(A)-Q(A).
\end{align}
Substituting this equality into \eqref{eq:vajda_tv_partition} yields
\begin{equation}
V=P(A)-Q(A)=Q(A^c)-P(A^c).
\label{eq:vajda_tv_event}
\end{equation}

Introduce the auxiliary probability-mass parameter
\[
\beta:=Q(A).
\]
By \eqref{eq:vajda_tv_event}, the probability masses of the two distributions on the binary partition $\{A,A^c\}$ are
\begin{align}
P(A)&=\beta+V,
&
Q(A)&=\beta,
\label{eq:vajda_binary_A}\\
P(A^c)&=1-\beta-V,
&
Q(A^c)&=1-\beta.
\label{eq:vajda_binary_Ac}
\end{align}
Non-negativity of these probabilities requires
\[
\beta\in[0,1-V].
\]

We next partition the $f$-divergence according to $A$ and $A^c$. Recall that
\[
D_f(P\parallel Q)
=\int_{\mathcal{X}}q(x)
f\left(\frac{p(x)}{q(x)}\right)\,dx.
\]
Thus,
\begin{align}
D_f(P\parallel Q)
&=\int_Aq(x)f\left(\frac{p(x)}{q(x)}\right)\,dx
+\int_{A^c}q(x)f\left(\frac{p(x)}{q(x)}\right)\,dx.
\label{eq:vajda_partition_divergence}
\end{align}

Consider first the integral over $A$. For $\beta=Q(A)>0$, we can rewrite it as
\begin{align}
\int_Aq(x)f\left(\frac{p(x)}{q(x)}\right)\,dx
&=Q(A)\int_A\frac{q(x)}{Q(A)}
f\left(\frac{p(x)}{q(x)}\right)\,dx.
\end{align}
The normalized density $q(x)/Q(A)$ defines the conditional distribution $Q(\cdot\mid A)$. Since $f$ is convex, Jensen's inequality gives
\begin{align}
\int_Aq(x)f\left(\frac{p(x)}{q(x)}\right)\,dx
&\geq
Q(A)f\left(
\int_A\frac{q(x)}{Q(A)}
\frac{p(x)}{q(x)}\,dx
\right) \notag\\
&=
Q(A)f\left(
\frac{P(A)}{Q(A)}
\right).
\label{eq:vajda_jensen_A}
\end{align}
Similarly, applying Jensen's inequality to the complementary set $A^c$ yields
\begin{align}
\int_{A^c}q(x)f\left(\frac{p(x)}{q(x)}\right)\,dx
&\geq
Q(A^c)f\left(
\frac{P(A^c)}{Q(A^c)}
\right).
\label{eq:vajda_jensen_Ac}
\end{align}

Combining \eqref{eq:vajda_partition_divergence},
\eqref{eq:vajda_jensen_A}, and \eqref{eq:vajda_jensen_Ac}, we obtain
\begin{equation}
D_f(P\parallel Q)
\geq
Q(A)f\left(\frac{P(A)}{Q(A)}\right)
+
Q(A^c)f\left(\frac{P(A^c)}{Q(A^c)}\right).
\label{eq:vajda_binary_bound}
\end{equation}
This inequality can be interpreted as the data processing inequality for the binary mapping
$T(x)=\mathbf{1}_{A}(x)$: aggregating the original sample space into the two events $A$ and $A^c$ cannot increase the $f$-divergence.

Substituting the probability masses in
\eqref{eq:vajda_binary_A} and \eqref{eq:vajda_binary_Ac} into
\eqref{eq:vajda_binary_bound}, we obtain
\begin{align}
D_f(P\parallel Q)
\geq
\beta f\left(\frac{\beta+V}{\beta}\right) \notag+(1-\beta)f\left(
\frac{1-\beta-V}{1-\beta}
\right).
\end{align}
Rearranging the arguments of $f$ gives
\begin{align}
D_f(P\parallel Q)
\geq
\beta f\left(1+\frac{V}{\beta}\right)
+(1-\beta)f\left(1-\frac{V}{1-\beta}\right).
\label{eq:vajda_parametric_lower_bound}
\end{align}

Since $\beta$ is an auxiliary parameter determined by the particular pair $(P,Q)$, while constrained only by $\beta\in[0,1-V]$, taking the infimum over all feasible values of $\beta$ yields
\begin{equation}
D_f(P\parallel Q)
\geq
\inf_{\beta\in[0,1-V]}
\left[
\beta f\left(1+\frac{V}{\beta}\right)
+
(1-\beta)f\left(1-\frac{V}{1-\beta}\right)
\right].
\end{equation}
Therefore,
\[
D_f(P\parallel Q)\geq L_f(V).
\]

The boundary cases $\beta=0$ and $\beta=1-V$ are understood by continuity through the corresponding limiting binary distributions. In particular, when a denominator vanishes, the expression is interpreted using the extended-valued convention for $f$-divergences.

Finally, the lower bound is tight. For any $V\in[0,1]$ and $\beta\in[0,1-V]$, consider the binary distributions
\[
P_\beta=(\beta+V,1-\beta-V),
\qquad
Q_\beta=(\beta,1-\beta).
\]
They satisfy
\[
D_{\mathrm{TV}}(P_\beta\parallel Q_\beta)
=\frac{1}{2}\left(
|\beta+V-\beta|
+
|1-\beta-V-(1-\beta)|
\right)
=V.
\]
Moreover,
\begin{align}
D_f(P_\beta\parallel Q_\beta)
&=
\beta f\left(\frac{\beta+V}{\beta}\right)
+
(1-\beta)f\left(
\frac{1-\beta-V}{1-\beta}
\right) \notag\\
&=
\beta f\left(1+\frac{V}{\beta}\right)
+
(1-\beta)f\left(1-\frac{V}{1-\beta}\right).
\end{align}
Thus, taking the infimum over $\beta\in[0,1-V]$ shows that $L_f(V)$ is achieved by binary distributions whenever the infimum is attained. More generally, a sequence of binary distributions can approach $L_f(V)$ arbitrarily closely.
Hence, no larger universal lower bound depending solely on $V=D_{\mathrm{TV}}(P\parallel Q)$ can hold for all probability distributions.

\paragraph{Derivation for the Optimal Constant.}

We next derive the optimal constant for a quadratic lower bound between
the total variation distance and an arbitrary $f$-divergence. Specifically,
we seek the largest constant $C_f>0$ such that
\begin{equation}
D_f(P\parallel Q)
\geq C_f D_{\mathrm{TV}}(P\parallel Q)^2
\end{equation}
holds uniformly for all probability distributions $P$ and $Q$.

Let
\[
V=D_{\mathrm{TV}}(P\parallel Q).
\]
From Vajda's tight lower bound established above, we have
\begin{equation}
D_f(P\parallel Q)\geq L_f(V).
\label{eq:quadratic_from_vajda}
\end{equation}
For any $V>0$, dividing both sides of \eqref{eq:quadratic_from_vajda}
by $V^2$ gives
\begin{equation}
\frac{D_f(P\parallel Q)}
{D_{\mathrm{TV}}(P\parallel Q)^2}
\geq
\frac{L_f(V)}{V^2}.
\label{eq:ratio_vajda_bound}
\end{equation}
Since the right-hand side depends only on the total variation distance,
a universal quadratic lower bound must hold for all possible values
$V\in(0,1]$. Therefore, taking the infimum over $V$ yields
\begin{equation}
C_f^{\mathrm{Vajda}}
:=
\inf_{0<V\leq1}
\frac{L_f(V)}{V^2}.
\label{eq:optimal_vajda_constant}
\end{equation}
By definition of the infimum,
\[
\frac{L_f(V)}{V^2}\geq C_f^{\mathrm{Vajda}}
\qquad\text{for all }V\in(0,1].
\]
Combining this inequality with \eqref{eq:quadratic_from_vajda}, we obtain
\begin{align}
D_f(P\parallel Q)
&\geq L_f(V) \notag\\
&\geq C_f^{\mathrm{Vajda}}V^2 \notag\\
&=C_f^{\mathrm{Vajda}}
D_{\mathrm{TV}}(P\parallel Q)^2.
\end{align}
Equivalently,
\begin{equation}
D_{\mathrm{TV}}(P\parallel Q)^2
\leq
\frac{1}{C_f^{\mathrm{Vajda}}}
D_f(P\parallel Q).
\label{eq:vajda_quadratic_final}
\end{equation}

It remains to show that $C_f^{\mathrm{Vajda}}$ is the largest possible
universal constant. Suppose there exists a constant $C>C_f^{\mathrm{Vajda}}$
such that
\[
D_f(P\parallel Q)\geq C D_{\mathrm{TV}}(P\parallel Q)^2
\]
holds for all probability distributions $P$ and $Q$. By the tightness of
Vajda's bound, for every $V\in(0,1]$ there exist binary distributions
$(P_\beta,Q_\beta)$ satisfying
\[
D_{\mathrm{TV}}(P_\beta,Q_\beta)=V
\]
and
\[
D_f(P_\beta\parallel Q_\beta)=L_f(V),
\]
where $\beta$ minimizes the binary lower-bound expression (or approaches
its infimum arbitrarily closely). Applying the assumed quadratic bound to
these distributions gives
\[
L_f(V)\geq CV^2,
\]
and hence
\[
\frac{L_f(V)}{V^2}\geq C
\qquad\text{for all }V\in(0,1].
\]
Taking the infimum over $V$ contradicts the definition of
$C_f^{\mathrm{Vajda}}$, since it would imply
\[
C_f^{\mathrm{Vajda}}
=\inf_{0<V\leq1}\frac{L_f(V)}{V^2}
\geq C>C_f^{\mathrm{Vajda}}.
\]
Therefore, no constant larger than $C_f^{\mathrm{Vajda}}$ can satisfy the
quadratic lower bound uniformly over all distributions. Consequently,
$C_f^{\mathrm{Vajda}}$ is the optimal universal quadratic constant.

\paragraph{Local Quadratic Behavior of the Vajda Bound.}
We further characterize the local quadratic behavior of the binary lower
bound around $V=0$. For a fixed $\beta\in(0,1)$, define
\begin{equation}
g_\beta(V)
=
\beta f\left(1+\frac{V}{\beta}\right)
+
(1-\beta)f\left(1-\frac{V}{1-\beta}\right).
\end{equation}
The Vajda bound gives
\begin{equation}
D_f(P\parallel Q)\geq g_\beta(V),
\qquad
V=D_{\mathrm{TV}}(P,Q).
\end{equation}

Since $f(1)=0$, we have
\begin{equation}
g_\beta(0)=0.
\end{equation}
Differentiating with respect to $V$ gives
\begin{align}
g_\beta'(V)
&=
f'\left(1+\frac{V}{\beta}\right)
-
f'\left(1-\frac{V}{1-\beta}\right),
\\
g_\beta''(V)
&=
\frac{1}{\beta}
f''\left(1+\frac{V}{\beta}\right)
+
\frac{1}{1-\beta}
f''\left(1-\frac{V}{1-\beta}\right).
\end{align}
Consequently,
\begin{equation}
g_\beta'(0)=0,
\qquad
g_\beta''(0)
=
\frac{f''(1)}{\beta(1-\beta)}.
\end{equation}

Applying Taylor's theorem around $V=0$, we obtain
\begin{align}
g_\beta(V)
&=
\frac{f''(1)}{2\beta(1-\beta)}V^2
+o(V^2),
\qquad V\to0.
\end{align}
Therefore, for every fixed $\beta\in(0,1)$,
\begin{equation}
D_f(P\parallel Q)
\geq
\frac{f''(1)}{2\beta(1-\beta)}
D_{\mathrm{TV}}(P,Q)^2
+o\left(D_{\mathrm{TV}}(P,Q)^2\right).
\end{equation}

Since
\[
\beta(1-\beta)\leq\frac14,
\]
the smallest possible local quadratic coefficient over
$\beta\in(0,1)$ is
\begin{equation}
C_f^{\mathrm{local}}
=
\min_{\beta\in(0,1)}
\frac{f''(1)}{2\beta(1-\beta)}
=
2f''(1),
\end{equation}
attained at $\beta=1/2$. This local coefficient should be distinguished
from the global Vajda constant
\begin{equation}
C_f^{\mathrm{Vajda}}
=
\inf_{0<V\leq1}\frac{L_f(V)}{V^2}.
\end{equation}
The former characterizes the local curvature of the binary lower bound
around $V=0$, whereas the latter controls the entire range
$V\in(0,1]$. Consequently,
\begin{equation}
C_f^{\mathrm{Vajda}}
\leq
C_f^{\mathrm{local}}
=
2f''(1),
\end{equation}
with equality when the global minimum of $L_f(V)/V^2$ is attained
asymptotically at $V=0$.

\section{Variance Reduction of Atomic Estimation}

We next analyze the variance reduction induced by Atomic Estimation.
Let
\begin{equation}
h(a)=f(r(s,a)),
\qquad a\sim\mu(\cdot\mid s),
\end{equation}
so that
\begin{equation}
D_f(\pi\parallel\mu)
=
\sum_a\mu(a\mid s)h(a).
\end{equation}

Since the variance of a random contribution decomposes into its second
moment minus its squared mean, we focus on the second-moment term, which
captures the contribution of large-magnitude samples and is the dominant
source of variance in long-tailed importance-weighted estimation.

Under MC estimation, the sampled divergence contribution is
\begin{equation}
\widehat D_f^{\mathrm{MC}}=h(a).
\end{equation}
Its second moment is
\begin{equation}
\mathbb{E}_{a\sim\mu}
\left[
\left(\widehat D_f^{\mathrm{MC}}\right)^2
\right]
=
\sum_a\mu(a\mid s)h(a)^2.
\end{equation}
Thus, the contribution of action $a$ to the second moment is
\begin{equation}
v_{\mathrm{MC}}(a)
=
\mu(a\mid s)h(a)^2.
\end{equation}

Atomic Estimation instead evaluates the probability-weighted atomic
contribution
\begin{equation}
\widehat D_f^{\mathrm{Atom}}
=
\mu(a\mid s)h(a).
\end{equation}
Its corresponding second moment is
\begin{equation}
\mathbb{E}_{a\sim\mu}
\left[
\left(\widehat D_f^{\mathrm{Atom}}\right)^2
\right]
=
\sum_a\mu(a\mid s)^3h(a)^2,
\end{equation}
so the contribution of action $a$ becomes
\begin{equation}
v_{\mathrm{Atom}}(a)
=
\mu(a\mid s)^3h(a)^2.
\end{equation}
Consequently,
\begin{equation}
v_{\mathrm{Atom}}(a)
=
\mu(a\mid s)^2v_{\mathrm{MC}}(a)
\leq
v_{\mathrm{MC}}(a),
\end{equation}
where the inequality follows from $\mu(a\mid s)\leq1$.

The suppression is particularly strong for rare actions. For
$\mu(a\mid s)\ll1$, the second-moment contribution of action $a$ is reduced
by the additional factor $\mu(a\mid s)^2$ under Atomic Estimation. Hence,
even when a rare action exhibits a large ratio deviation and consequently
a large raw value of $h(a)$, its contribution to the probability-weighted
proximal constraint remains controlled.

This analysis shows that Atomic Estimation mitigates the long-tail
sensitivity of MC-based constraints by suppressing the second-moment
contribution of low-probability actions. It therefore provides a
probability-aware mechanism for improving the stability of proximal
estimation, while distinguishing the magnitude of a ratio deviation from
its actual contribution to the overall divergence.


\end{document}